\documentclass[runningheads]{llncs}
\usepackage[T1]{fontenc}
\usepackage{graphicx}
\graphicspath{ {./figures/} }
\begin{document}
\title{Enhancing Ultra-low-field MRI with Segmentation-guided Adversarial Learning}
\titlerunning{Segmentation-Guided Adversarial Learning}
\begingroup
\renewcommand\thefootnote{}
\footnotetext{Preprint of the submitted version of a paper to appear in: \emph{Enhancing Ultra-Low-Field MRI with Paired High-Field MRI Comparisons for Brain Imaging}, Lecture Notes in Computer Science (LNCS, volume 16293), Springer, 2026. The final publication will be available online via \url{https://link.springer.com/book/9783032233431}.}
\endgroup
\authorrunning{J. Grover et al.}
%
\institute{Anonymised Affiliation \email{email@anonymised.com}}
\author{James Grover\inst{1} \and Andrew Phair\inst{1} \and Michael Ferraro\inst{1} \and David E.J. Waddington\inst{1}}
\authorrunning{J. Grover et al.}
%
\institute{Image X Institute, Sydney School of Health Sciences, Faculty of Medicine and Health, The University of Sydney \\ Correspondence: \email{james.grover@sydney.edu.au}}

\maketitle              
\begin{abstract}
Ultra-low-field (ULF) MRI offers portable and low-cost imaging but suffers from poor image quality. To address this, we present our submission to the 2025 ULF Enhancement Challenge (ULF-EnC), where the goal is to synthesise high-field-like MRIs from 64~mT scans. Our pipeline enhances ULF MRI through a combination of anatomical conditioning and model ensembling. We first generate tissue segmentation priors using a Swin UNETR trained solely on challenge-provided data. These priors condition two independent enhancement networks—a CycleGAN and a transformer-based residual enhancement model (T-REX)—each trained to synthesise 3~T-like MRIs. Outputs from both models are combined using a weighted average. Our approach produces enhanced MRIs that were comparable to high-field scans both quantitatively and qualitatively.

\end{abstract}
\section{Introduction}

Magnetic resonance imaging (MRI) provides unparalleled structural and functional insight into human anatomy, yet remains inaccessible to much of the global population due to two major limitations: cost and portability. High-field (HF) MRI systems (e.g., 3~T) are expensive, stationary, and require specialised infrastructure, making them largely unavailable in low-and middle-income countries and impractical for critically ill or immobile patients \cite{kwikima2024looking}. In response, portable ultra-low-field (ULF) MRI has emerged as a promising alternative, enabling bedside and community-based imaging with lower infrastructure demands \cite{obungoloch2023site,turpin2020portable,sarracanie2015low}. These systems operate at much lower magnetic field strengths (e.g., 64~mT), leading to significantly reduced signal-to-noise ratio (SNR), suboptimal image contrast, and susceptibility to artefacts from limited RF shielding and gradient performance. Addressing these limitations to unlock the diagnostic potential of ULF MRI remains a critical challenge for equitable medical imaging.

Recent research has focused on applying machine learning to enhance ULF MRI by translating low-SNR ULF scans into HF-quality images \cite{shimron2024accelerating}. Most of these approaches leverage paired datasets with subjects that have been imaged with both ULF and conventional HF MRI, enabling supervised learning and evaluation. Since radiologists and clinicians are already familiar with the contrast and resolution characteristics of 3-T MRI, synthesising images in this domain can aid interpretation and increase clinical confidence.  The Ultra‑Low‑Field MRI Image Enhancement Challenge (ULF‑EnC) \cite{ulfenczenodo} - consists of paired 64-mT and 3-T brain MRI data across multiple contrasts (T1, T2, FLAIR).

Paired image-translation approaches have demonstrated that generative models can lift 64-mT scans toward 3 T appearance, improving visual quality and enabling standard neuro-morphometry analyses \cite{islam2023improving,lucas2025multisequence,iglesias2022quantitative,gopinath2025low}. However, existing enhancement pipelines rarely condition synthesis on explicit tissue logits or combine complementary generators within the stringent data constraints of portable ULF MRI—limitations that the present study seeks to overcome.

Here we demonstrate a structured deep learning pipeline for enhancing ULF (64~mT) MRI using paired data from the ULF-EnC challenge. Our approach is guided by the principle that any successful enhancement method must ultimately infer and preserve underlying brain anatomy. To this end, we adopt a segmentation-driven strategy: anatomical priors are extracted via Swin UNETR logits trained to predict HF tissue maps generated using automated FSL tools (i.e. using no external data). The resulting logits are used to condition two complementary enhancement networks—a CycleGAN and a transformer-based residual enhancement model (T-REX). Their outputs are combined using a weighted average to produce a final enhancement. We hypothesised that embedding anatomical structure into the enhancement process is essential for diagnostic reliability. Additionally, by combining the enhancements from independently trained models image quality and robustness could be increased over single model approaches.

\section{Methods}
\subsection{Data pre-processing}
The ULF-EnC training dataset is comprised of 50 subjects, each containing ULF (64 mT, \textit{Hyperfine} Swoop) and HF (3 T, \textit{Siemens} Biograph mMR) brain MRIs with T1, T2, and FLAIR contrast \cite{islam2023improving,dayarathna2024deep,dayarathna2024ultra,dayarathna2025mccad}. For algorithm development, we assigned 45 subjects as a train dataset and five subjects as an internal validation dataset to guide model development (c.f., the challenge’s external validation set). 

\subsection{Overall pipeline}
\begin{figure}
    \centering
    \includegraphics[width=1.00\linewidth]{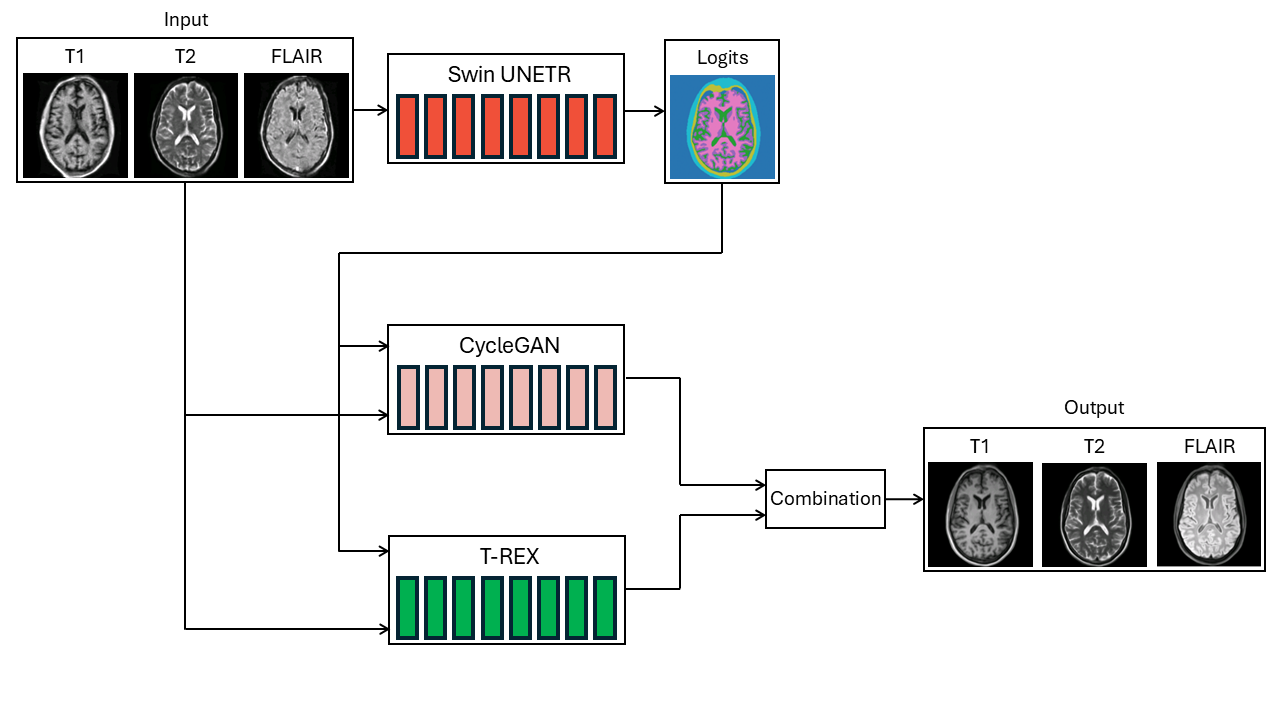}
    \caption{Overall pipeline for producing enhanced ultra-low-field MRIs. Input ultra-low-field MRIs are enhanced by individual CycleGAN and T-REX networks with additional segmentation priors by Swin UNETR. Enhancements are combined to produce final output MRIs.}
    \label{fig:overview}
\end{figure}

We developed a pipeline for enhancing 64 mT ULF multi-contrast inputs (Fig. \ref{fig:overview}). The pipeline consists of four phases:
\begin{enumerate}
    \item Segmentation prediction using Swin UNETR,
    \item ULF enhancement using CycleGAN,
    \item ULF enhancement using T-REX, and
    \item Enhancement combination.
\end{enumerate}

All models were trained and implemented using the PyTorch deep learning framework \cite{paszke2019pytorch}.

\subsection{Segmentation prediction using Swin UNETR}
Pre-trained neural networks can improve the performance of downstream neural networks by injecting additional priors. We hypothesised that by training a segmentation network to generate segmentations of various anatomical structures from the ULF MRI, deep features could be transferred to the task of enhancement.

To generate ground truth segmentations for training, we developed a multi-step preprocessing pipeline applied to the 3 T T1- and T2-weighted images. Importantly, caution was taken to ensure that no external data or pre-trained models were used at any stage of this segmentation process—only the challenge-provided data and publicly available, non-learning-based and non-atlas-based tools were used.

FSL’s Brain Extraction Tool (BET) was applied to the 3 T T1- and T2-weighted images jointly, producing brain, skull, and scalp masks \cite{jenkinson2012fsl,smith2002fast,jenkinson2005bet2}. BET employs a deformable surface model initialised as a sphere, which expands iteratively to fit the brain boundary by optimising intensity contrast and surface smoothness.

FSL’s FAST was then used to segment the T1-weighted brain image into cerebrospinal fluid (CSF), grey matter (GM), and white matter (WM) \cite{zhang2002segmentation}. FAST uses a hidden Markov random field (HMRF) model with expectation-maximisation to simultaneously perform bias field correction and tissue classification.

The resulting masks were combined into a six-class labelmap: background, CSF, GM, WM, skull, and scalp. All voxels not assigned to one of these classes were labelled as background. A Swin UNETR model was trained to predict the six-class 3 T segmentation directly from the three 64 mT contrasts (T1, T2, and FLAIR) \cite{hatamizadeh2021swin}. The network architecture used 3D Swin Transformer blocks with hierarchical encoding, and was implemented using the MONAI framework \cite{liu2021swin,cardoso2022monai}. The number of trainable parameters of this model was $\approx$ 62M. 

The loss function was a combination of multi-class Dice loss and cross-entropy loss. Data augmentation was applied using MONAI’s 3D transform library, including random affine transformations, flipping, and intensity perturbations. In the final training phase, augmentation was disabled and training continued on unaltered data. Model performance was evaluated using the mean Dice similarity coefficient on a separate validation set. Segmentation model weights were frozen after this training for all subsequent use.

\subsection{ULF enhancement using CycleGAN}
The image-to-image translation framework CycleGAN was adapted to incorporate 3D network architectures and trained to perform the task of low-field-to-high-field enhancement across three image contrasts simultaneously \cite{zhu2017unpaired}.

The principle underlying the CycleGAN approach is that by simultaneously training two generator-discriminator network pairs operating in opposite directions, the conditional generators learn to generate images that belong within the target output distribution while also relating to the input image. To enforce this, cycle losses in each domain are calculated between input images and the output that is yielded by applying the two generators successively. 

The architecture of the generator network \cite{he2016deep} was adapted from the 9-block residual network included in the publicly available CycleGAN code. We note that this 2D generator architecture has previously been successfully applied to an ULF MRI enhancement task \cite{islam2023improving}. To provide greater volumetric context, 2D convolution operations were replaced with their 3D equivalent and, to reduce chequerboard artefacts, transpose convolution operations were replaced with trilinear upsampling followed by 3D convolutions.

We experimented with methods of performing segmentation conditioning for the challenging ULF to 3 T MRI enhancement task. Initially, this was achieved by concatenating segmentation probability channels (produced by the pre-trained Swin UNETR) at input into the first layer, resulting in the ULF-to-HF generator receiving a nine-channel volume (three contrast-weighted images + six segmentation probability channels) and outputting an enhanced three-channel image volume). Second, as an alternative to concatenation, we employed spatially adaptive normalisation to enhance the inductive bias provided by semantic conditioning, replacing all normalisation layers in the ULF-to-HF generator with SPADE \cite{park2019semantic}. To maintain parameter parity, and due to fewer input channels, we correspondingly reduced the number of bottleneck channels in the SPADE ULF-to-HF generator, giving a resultant parameter count of $\approx 33$M parameters for each generator architecture.

In each of these cases, the HF-to-ULF generator received only the three-channel HF image volumes as input, and output corresponding synthesised three-channel low-field image volumes.

The two corresponding discriminator networks were trained to distinguish between genuine and generated images in each domain.

An additional paired loss term was introduced between the HF and enhanced low-field 3D image slabs, calculated as the negative of the weighted challenge score (\ref{eqn:weighting}). During the course of training this loss was modulated by a penalty weighting factor proportional to the arctangent of the epoch number, with the intention of forcing the generator to first learn to produce enhanced images consistent with the HF image distribution, and then to fine-tune these enhanced images to optimise the challenge metrics.

Due to both memory and data scarcity constraints, the CycleGAN was trained on 40-slice slabs randomly selected from the total 3D image volumes each epoch. At inference, overlapping slabs (with a stride of 5) were enhanced and then averaged to form a composite enhanced volume.

\subsection{ULF enhancement using T-REX}
We developed a 3D volume-to-volume transformer residual hybrid network (T-REX) for image synthesis. T-REX is comprised of an encoding arm, decoding arm, and information bottleneck utilising skip connections between encoder and decoder (similar to a U-NET) \cite{ronneberger2015u}. The encoding arm consists of convolutional layers and downsampling layers via strided convolution. The decoding arm consists of convolutional layers and upsampling layers via interpolation. At the information bottleneck, the latent tensor is transformer to a linear representation and processed by a transformer encoder \cite{vaswani2017attention}. This processed latent tensor is then reshaped back into the image domain and provided to the decoding arm. Skip connections between stages of the encoding and decoding arm incorporate residual blocks based off the enhanced deep super-resolution (EDSR) architecture \cite{lim2017enhanced,grover2024super}. Segmentation information (produced by the pre-trained Swin UNETR) were concatenated as six additional channels to the ULF MRIs. The number of trainable parameters for T-REX was $\approx$ 24M. 

Due to the significant distribution shift between ULF and HF, T-REX was trained adversarially using a conditional GAN (cGAN) framework \cite{isola2017image}. The generator loss included two terms, a discriminator term (based on least squares) and a content term \cite{mao2017least}. The content term ensures that generated enhancements closely resemble the HF MRIs on a pixel-wise basis; This was a weighted combination of L1 loss, peak signal to noise ratio (PSNR), and a Sobel filter-based loss (to directly optimise for high spatial frequency features). Once T-REX had converged under adversarial training ($\approx$ 100 epochs), further fine-tuning was conducted in a traditional supervised learning approach ($\approx$ 10 epochs).

\subsection{Enhancement combination}
Model ensembling is a powerful technique to ultimately boost performance and increase robustness in the context of medical imaging deep learning \cite{ferreira2024we}. By combining both CycleGAN and T-REX enhancements using a simple weighted average, a single enhancement could be produced from two independently trained networks. We optimised this weighting on the ULF EnC validation set.

\subsection{Measuring performance}
A qualitative and quantitative analysis was performed on the enhanced validation ULF MRIs from the ULF EnC challenge. Qualitative analysis included the inspection of enhancements with regards to overall image quality, hallucinated features, and artefacts. We compare each individual model in addition to the combination. HF references were not available for this specific analysis.

We use the quantitative metrics as measured by challenge organisers to ensure consistent reporting with the final hidden test set. Two sets of metrics were reported: one involving no use of masking and the other involving the use a background mask (to remove irrelevant features from metric calculation). We calculated the weighted scores by applying the formula as detailed by challenge organisers (\ref{eqn:weighting}).

\begin{equation}
    \mathrm{Weighted} = 0.7\cdot \mathrm{SSIM} + 0.1 \cdot \mathrm{PSNR} + 0.1 \cdot (1 - \mathrm{MAE}) + 0.1 \cdot (1 - \mathrm{NMSE}) \label{eqn:weighting}
\end{equation}

To observe hallucinations in our reconstructions we focused on regions that are unlikely to be captured by the point-of-care imaging system (i.e. ULF). The nose was not present in some of the ULF MRIs which was the basis for our hallucination analysis.

\section{Results}
CycleGAN, T-REX, and their combination were all capable of producing enhanced 3 T-like MRIs from ULF inputs. Quantitative metrics are provided in Table \ref{table:quantitative} as measured during the validation phase of the challenge. The combined approach increased masked SSIM and PSNR while decreasing error metrics.

\begin{table}[htbp]
\centering
\caption{Quantitative metrics. Values provided are the [unmasked|masked] score averaged for all contrasts and validation subjects. Best scores for each metric are \underline{underlined}.}
\begin{tabular}{ |p{2.1cm}|p{2.4cm}|p{2.4cm}|p{2.4cm}|}
 \hline
  & \hfil \hfil \textbf{CycleGAN} & \hfil \textbf{T-REX} & \hfil \textbf{Combined} \\
  \hline
  \hline
  \textbf{SSIM} ($\uparrow$) & \hfil \underline{0.839} | 0.705 & \hfil 0.816 | 0.684  & \hfil 0.832 | \underline{0.711} \\ 
    \hline
  \textbf{PSNR} ($\uparrow$) & \hfil 23.154 | 29.559 & \hfil 22.956 | 29.601 & \hfil \underline{23.372} | \underline{30.416}  \\ 
    \hline
  \textbf{MAE} ($\downarrow$) & \hfil \underline{0.031} | 0.068 & \hfil 0.033 | 0.071 & \hfil \underline{0.031} | \underline{0.067} \\ 
    \hline
  \textbf{NMSE} ($\downarrow$) & \hfil 0.065 | 0.148 & \hfil 0.068 | 0.155  & \hfil \underline{0.063} | \underline{0.141}  \\ 
    \hline
  \textbf{Weighted} ($\uparrow$) & \hfil 3.093 | 3.628 & \hfil 3.057 | 3.616  & \hfil \underline{3.110} | \underline{3.719}  \\ 
 \hline
\end{tabular}
\label{table:quantitative}
\end{table}

\begin{figure}
    \centering
    \includegraphics[width=1\linewidth]{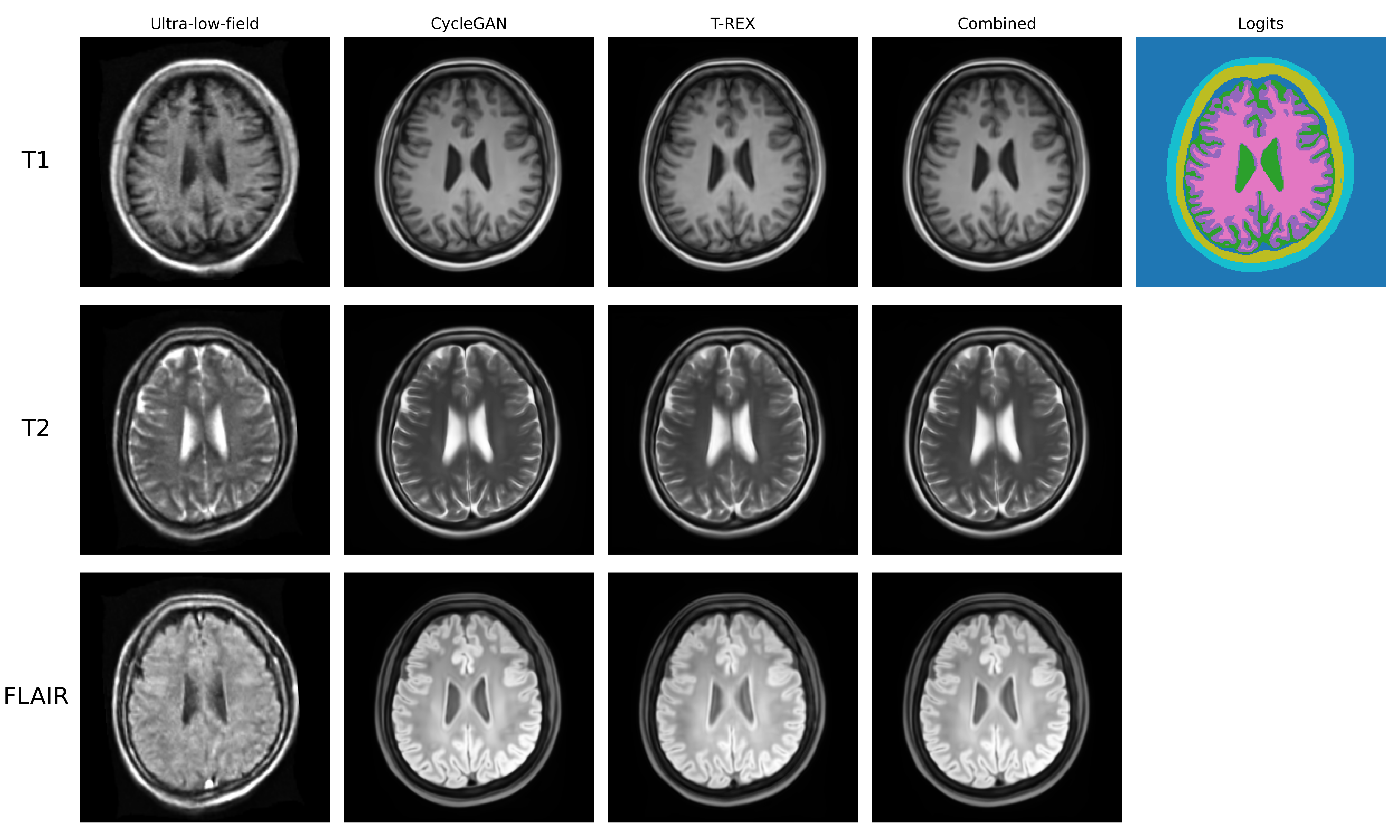}
    \caption{Enhancement demonstration from an axial slice. All enhancement methods were capable of producing synthetic 3 T-like enhancements. Logits were used to guide enhancement models.}
    \label{fig:brain_results}
\end{figure}

Both CycleGAN and T-REX were capable of independently producing high quality enhancements that could be combined effectively (Fig. \ref{fig:brain_results}).

\begin{figure}
    \centering
    \includegraphics[width=1\linewidth]{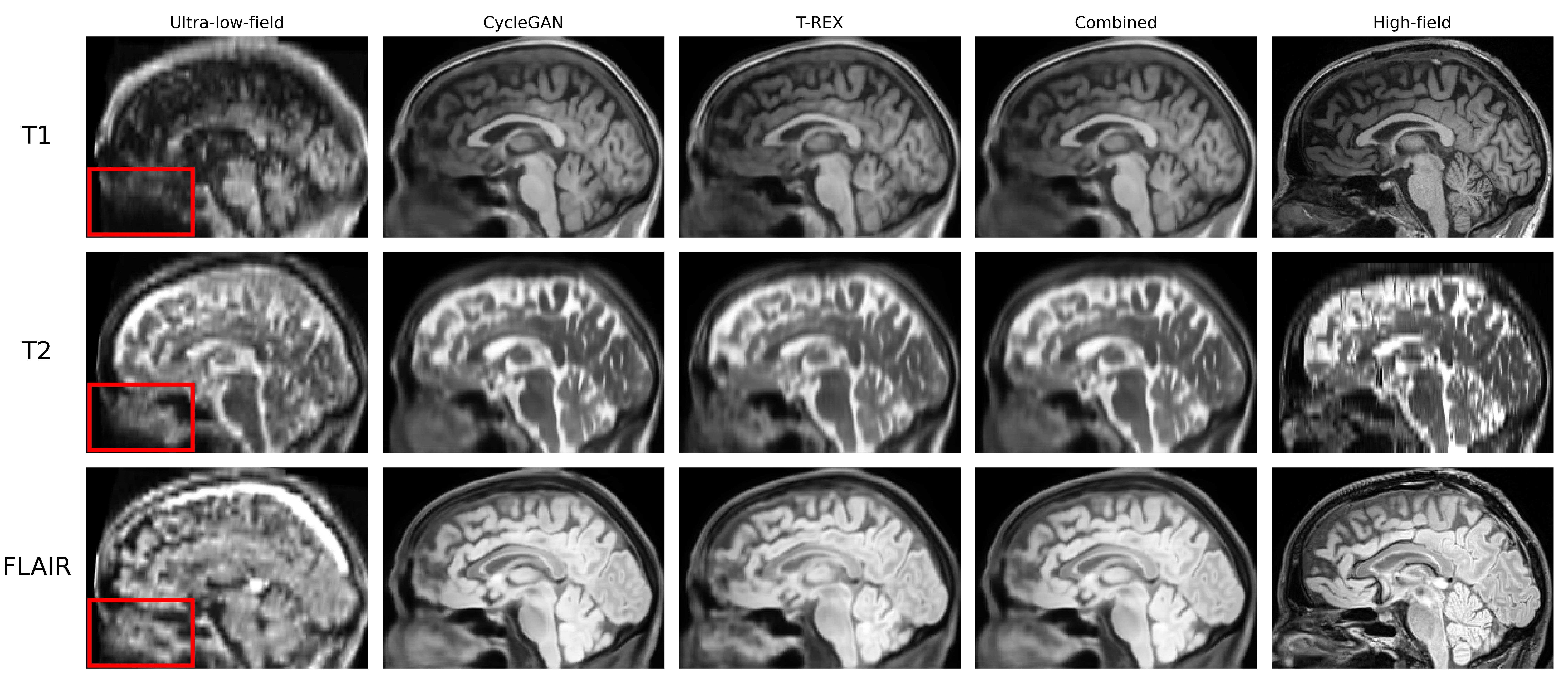}
    \caption{Hallucination analysis from a sagittal slice. A clear signal void is present around the nose region in the ultra-low-field. Enhancement methods filled in these areas with features highlighting the need for detection and mitigation of hallucinations in ULF enhancement models.}
    \label{fig:brain_results_hallucination}
\end{figure}

The results for our hallucination experiment are presented in (Fig. \ref{fig:brain_results_hallucination}). Hallucinated features, namely surrounding the nose, were present in the enhancement models. These are likely due to the optimisation process rewarding best guesses in regions of low signal to bolster full image reference metrics such as PSNR.

\section{Discussion}
Our team produced a comprehensive enhancement pipeline for generating enhanced ULF MRIs. We focus our discussion here on two major points: hallucinations and metric optimisation.
\subsection{Hallucinations}
In the context of this challenge, hallucinations could be anatomical features that are generated by an enhancement model that do not exist and are unable to be inferred from the ULF. These could have a catastrophic impact on patient's care due to misdiagnoses. We observed hallucinations in the nose region (Fig. \ref{fig:brain_results_hallucination}) in the presence of a signal void (most apparent in the subject's T1 MRI). While the region of interest is usually constrained to the brain for ULF, our results here underpin the importance of identifying and limiting these hallucinations before widespread deployment of these models.

\subsection{Metric optimisation}
A notable challenge encountered during this work relates to Goodhart’s Law, which states that “when a measure becomes a target, it ceases to be a good measure.” This became evident in our efforts to optimise full-volume metrics such as PSNR and SSIM. The ULF scanner has markedly reduced receive sensitivity in regions such as the mouth and nose, leading to a genuine absence of signal in these areas. However, because these regions remain present in the paired 3 T reference, scalar metrics implicitly penalise any model that fails to reconstruct them. As a result, optimisation often favoured models that smoothed or blurred these regions—despite having no supporting signal—over models that more faithfully preserved intracranial structures. In several cases, models that produced visually sharper and more anatomically plausible brain reconstructions were outperformed (in PSNR) by blurrier models that filled in extracranial areas with low-confidence estimates. This misalignment illustrates how reliance on global image similarity metrics can drive development away from clinically relevant goals. To mitigate this, we recommend adopting region-of-interest metrics that focus on the brain, and incorporating downstream task-specific evaluations such as segmentation accuracy or tissue volume agreement. These provide a more meaningful assessment of image fidelity in regions where the data are actually informative and clinically useful.

\section{Conclusion}
We present our submission to the ultra-low-field enhancement (ULF EnC) challenge. Our anatomy-driven ensemble pipeline was capable of generating 3 T-like MRIs from 64 mT inputs of multiple contrasts (T1, T2, FLAIR).

\subsubsection{\ackname} J.G. is supported by an Australian Government Research Training Program scholarship and an Australian National Health and Medical Research Council (NHMRC) Investigator Grant supplementary scholarship. D.E.J.W. is supported by an Australian National Health and Medical Research Council (NHMRC) Investigator Grant (2017140).

\bibliographystyle{splncs04}
\bibliography{references}
\end{document}